\documentclass[letterpaper]{article} 
\usepackage{aaai25}  
\usepackage{times}  
\usepackage{helvet}  
\usepackage{courier}  
\usepackage[hyphens]{url}  
\usepackage{graphicx} 
\usepackage{natbib}  
\usepackage{caption} 
\usepackage{algorithm}
\usepackage{algorithmic}

\usepackage{xcolor}
\usepackage{multicol}
\usepackage{multirow}

\usepackage{amsmath}
\usepackage{amssymb}

\usepackage[misc]{ifsym}
\usepackage{utfsym}

\usepackage{newfloat}
\usepackage{listings}
\DeclareCaptionStyle{ruled}{labelfont=normalfont,labelsep=colon,strut=off} 
\floatstyle{ruled}
\newfloat{listing}{tb}{lst}{}
\floatname{listing}{Listing}
\title{DGFamba: Learning Flow Factorized State Space for \\ Visual Domain Generalization}

\author{
    Qi Bi\textsuperscript{\rm 1}\equalcontrib,
    Jingjun Yi\textsuperscript{\rm 1}\equalcontrib, Hao Zheng\textsuperscript{\rm 1}\footnotemark[2], 
    Haolan Zhan\textsuperscript{\rm 2},
    Wei Ji\textsuperscript{\rm 3}, Yawen Huang\textsuperscript{\rm 1}, Yuexiang Li\textsuperscript{\rm 4}\footnotemark[2]
    }
\affiliations{
    \textsuperscript{\rm 1}Jarvis Research Center, Tencent YouTu Lab, ShenZhen, China \\
    \textsuperscript{\rm 2}Faculty of Information Technology, Monash University, Melbourne, Australia \\
    \textsuperscript{\rm 3}School of Medicine, Yale University, New Haven, United States \\
    \textsuperscript{\rm 4}Faculty of Science and Technology, University of Macau, Macau \\

    \{q\_bi, rsjingjuny\}@whu.edu.cn, howzheng@tencent.com, yuexiang.li@ieee.org
}

\usepackage{bibentry}

\begin{document}

\maketitle

\begin{abstract}
Domain generalization aims to learn a representation from the source domain, which can be generalized to arbitrary unseen target domains. 
A fundamental challenge for visual domain generalization is the domain gap caused by the dramatic style variation whereas the image content is stable.
The realm of selective state space, exemplified by VMamba, demonstrates its global receptive field in representing the content.
However, the way exploiting the domain-invariant property for selective state space is rarely explored.
In this paper, we propose a novel Flow Factorized State Space model, dubbed as DGFamba, for visual domain generalization.
To maintain domain consistency, we innovatively map the style-augmented and the original state embeddings by flow factorization.
In this latent flow space, each state embedding from a certain style is specified by a latent probability path.
By aligning these probability paths in the latent space, the state embeddings are able to represent the same content distribution regardless of the style differences.
Extensive experiments conducted on various visual domain generalization settings show its state-of-the-art performance. 
\end{abstract}

%

\section{Introduction}
\label{sec1}

In many real-world scenarios, the distributions between the source and target domains are not independently and identically distributed (i.i.d).
Visual domain generalization handles the domain shift. 
It learns an image representation extracted from the source domain images, and aims to generalize to arbitrary unseen target domains \cite{perry2022causal,fang2020rethinking,hendrycks2021many,geirhos2021partial}.
Its key challenge lies in the domain gap caused by the dramatic style variation whereas the cross-domain image content is stable.

Visual domain generalization has been extensively studied in the past decade.
A variety of advanced machine learning techniques \cite{sagawa2019distributionally,krueger2021out,huang2020self,blanchard2021domain,zhou2024mixstyle,nam2021reducing} have been proposed to eliminate the impact of cross-domain styles.
However, these methods rely heavily on the convolutional neural network (CNN) \cite{he2016deep} as the image encoder, which has a limited local receptive field.
Since the local receptive field is more sensitive to the style variation and less expressive to the global-wise image content, modern domain generalization methods \cite{sultana2022self,li2023sparse} have shifted the image encoder from CNN to Vision Transformer (ViT) \cite{dosovitskiy2020image}, which is more capable to represent the image content owing to the self-attention mechanism.

\begin{figure}[!t]
  \centering
   \includegraphics[width=1.0\linewidth]{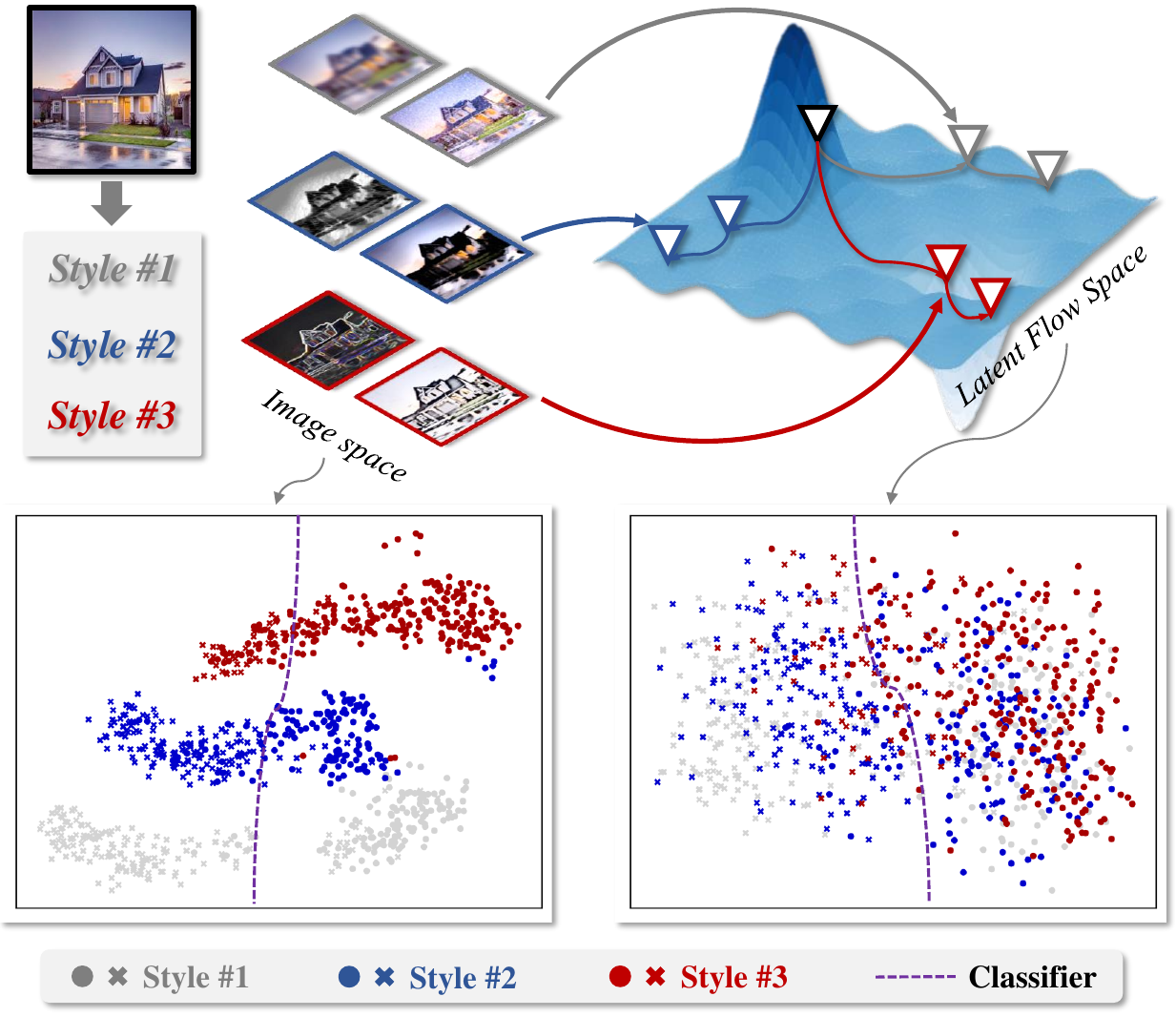}
   \caption{
   General idea of the proposed DGFamba. 
   The latent flow space aligns the probability path between the pre- and post- style augmented state embeddings to enhance the robustness to the style change.
   }
   \label{motivation}
\end{figure}

More recently, selective state space model (SSM), exemplified by VMamba \cite{liu2024vmamba} and Vision Mamba \cite{zhu2024vision}, has become the new paradigm for visual representation learning.
SSM converts the image into patch sequences and exploits the visual information from recurrent modeling, which demonstrates a more global receptive field to represent the content.
Such property provides a new feasible path for visual domain generalization, where a robust image content representation is critical.
However, the selective scanning is implemented in a fixed way regardless of the image features from different domains, which are usually distributed differently in the feature space.
It may not necessarily learn a consistent state embedding before and after the style augmentation.
Therefore, our question arises:
\textit{How to ensure that the selective state space is invariant to the cross-domain style shift?}

In this paper, we propose a flow factorized state space model, dubbed as DGFamba, for visual domain generalization.
Its general idea is to maintain the global receptive field of SSM to represent the content while at the same time empower SSM with style-invariant property.
Specifically, we introduce flow factorization \cite{song2023latent,song2024flow}, which maps the state embeddings between two styles by a probability path in the latent flow space (illustrated in Fig.~\ref{motivation}). 
By aligning the probability paths between the pre- and post- style augmented state embeddings, the selective state space is able to represent the same content distribution regardless of the style differences.

\begin{figure}[!t]
  \centering
   \includegraphics[width=1.0\linewidth]{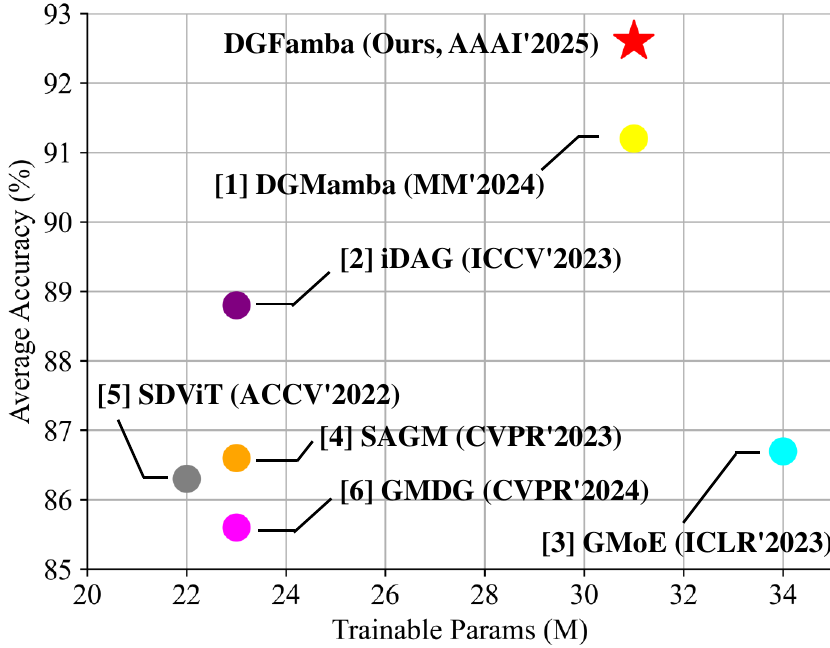}
   \caption{Performance on unseen domains (classification accuracy in \%) v.s. trainable parameter number (in million, M). The proposed DGFamba outperforms the state-of-the-art methods.
   }
   \label{teaser}
\end{figure}

The proposed DGFamba consists of three key components, namely, state style randomization, state flow encoding, and state flow constraint.
Specifically, before the selective scanning and recurrent modeling, the state style randomization maximizes the style diversity of the state embedding.
The style, parameterized by mean and standard deviation \cite{huang2017arbitrary}, is randomly sampled from a uniform distribution.
Then, the state flow encoding component projects the pre- and post- style hallucinated state embeddings into the latent flow space, and factorizes their latent probability path.
Finally, the state flow constraint aligns the latent probability path between the pre- and post- hallucinated state embeddings, so that the property of style invariant is achieved.

Our contributions can be summarized as follows.

\begin{itemize}
    \item We conduct an initial exploration of harnessing SSM for visual domain generalization, and propose a flow factorized state space modeling method (DGFamba).
    \item We introduce the flow factorization to represent the pre- and post- style hallucinated state embedding, which theoretically warrants the style invariant property of SSM.
    \item Experiments demonstrate that the proposed DGFamba not only significantly outperforms existing CNN and ViT based methods, but also surpasses the contemporary DGMamba by up to 1.5\% in top-1 accuracy.
\end{itemize}

\section{Related Work}
\label{Sec2}

\noindent \textbf{Mamba and Vision Mamba.}
Selective State Space Modeling (SSM), exemplified by Mamba and its variations~\cite{he2024mambaad,li2024videomamba,liu2024vmamba,wang2024mambaunet,xiao2023masked,bi2024samba}, is an emerging representation learning tool, which possesses global receptive fields with only linear complexity. 
In the field of computer vision,  
VMamba~\cite{liu2024vmamba} and Vim~\cite{zhou2021domain}
are pioneering works that adapt SSM for visual representation learning. 

\noindent \textbf{Domain Generalization.}
Most prior works use CNN as their backbone.
A variety of machine learning techniques, such as empirical risk minimization~\cite{xu2021fourier,huang2020self}, domain alignment~\cite{wang2023closer,nam2021reducing,wang2022variational,zhang2022delving}, domain augmentation~\cite{zhou2020learning,zhou2024mixstyle}, ensemble learning~\cite{kim2021selfreg,chen2022mix,chu2022dna}, frequency decoupling~\cite{bi2024generalized,yi2024learning,bi2024bwg,bi2024fada}, and meta learning~\cite{dou2019domain,du2020learning,zhao2021learning},
have been proposed. 
Vision Transformer
has demonstrated its superiority in visual domain generalization~\cite{noori2024tfs,zhang2022delving}.
Techniques such as mixture of experts~\cite{li2023sparse} and token-wise stylization~\cite{noori2024tfs}~have been studied.
More recently, \cite{long2024dgmamba}~made an earlier exploration to harness SSM for this task. However, the proposed DGMamba~\cite{long2024dgmamba}~only focused on improving the hidden space and patch embedding.
The style invariant property, which is crucial for visual domain generalization, remains unaddressed.

\noindent \textbf{Flow Factorization}
\cite{song2023latent,song2024flow}, as an emerging representation learning tool, holds a unique position to understand both 
disentangled and equivalent representations.
Inspired by the general idea that the representation distribution is encouraged to be factorial without substantially affecting the quality \cite{kim2018disentangling}, the probability of each transformation is modeled as a flow by the gradient field of some learned potentials following fluid mechanical dynamic optimal transport.
However, \textit{to the best of our knowledge}, flow factorization has so far rarely been explored in the context of domain generalization, especially for the style invariant properties.

\begin{figure*}[!t]
    \centering
    \includegraphics[width=1.00\linewidth]{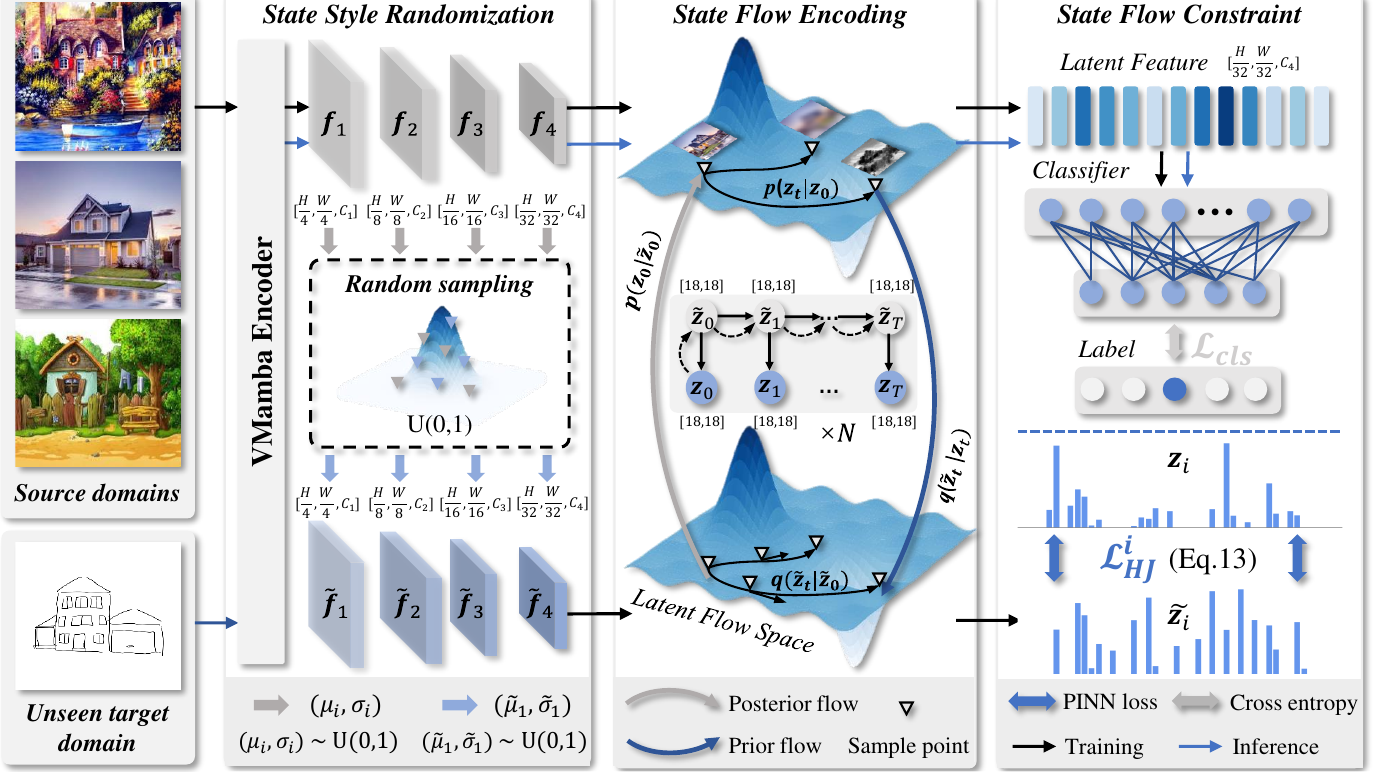}
    \vspace{-0.4cm}
    \caption{The proposed flow factorized state space model, dubbed as DGFamba, consists of three key components, namely, State Style Randomization, State Flow Encoding, and State Flow Constraint.}
    \label{framework}
\end{figure*}

\section{Methodology}
\label{Sec3}

Fig.~\ref{framework}~gives an overview of the proposed DGFamba, which consists of three key components, namely, state style randomization (SSR), State Flow Encoding (SFE), and State Flow Constraint (SFC), respectively. 

\subsection{State Style Randomization}
\label{sec4.1}

\textbf{Backbone \& State Embedding.} 
We use the VMamba~\cite{liu2024vmamba} as the backbone, assuming it consists of $N$ sequential layers from four feature blocks, denoted as $L_1$, $L_2$, $\cdots$, $L_N$.
For each layer, the output state embedding is denoted as $\boldsymbol{f}_i$.

\noindent \textbf{State Style Representing.} 
The state embedding $\boldsymbol{f}_i$ from a certain layer $L_i$ ($i=1,2, \cdots, N$) is supposed to generalize well to unseen target domains, where the dramatically different style variation leads to the distribution shift.
To realize this objective, the first step is to quantify and represent the style.
Following the common definition of style \cite{huang2017arbitrary}, the channel-wise mean 
$\boldsymbol{\mu}_{i}$ and standard deviation $\boldsymbol{\sigma}_{i}$ is used to quantize the style of $\boldsymbol{f}_i$, given by 
\begin{equation} 
\label{mean}
\boldsymbol \mu_{i} =  \frac{1}{C} \sum_{c=1}^{C} \boldsymbol{f}_{i}^c,
\quad 
\boldsymbol{\sigma}_{i} =  \sqrt{ \frac{1}{C} \sum_{c=1}^{C} (\boldsymbol{f}_{i}^c - \boldsymbol \mu_{i})^2}.
\end{equation}
where $C$ denotes the channel size.
After normalization, the per-channel mean and standard deviation value ranges from 0 to 1, denoted as $\mu_{i,m} \in \mathbb{R}^{[0,1]}$ and $\sigma_{i,m} \in \mathbb{R}^{[0,1]}$.

\noindent \textbf{Style Randomization.}
The model can learn the styles only from the source domain.
To enrich the style diversity, we hallucinate the state embedding $\boldsymbol{f}_{i}$ and generate the style augmented state embedding $\boldsymbol{\tilde{f}}_{i}$ with a random style.
Specifically, we randomize the hallucinated styles $[\boldsymbol {\widetilde{\mu}}_{i},\boldsymbol{\widetilde{\sigma}}_{i}]^\mathrm{T}$ from the entire style space $\mathcal{S} \subset \mathbb{R}^{[0,1] \times [0,1]}$:
\begin{equation} 
\label{sampling}
\widetilde{\mu}_{i,m} \sim [0,1], \quad \widetilde{\sigma}_{i,m} \sim [0,1].
\end{equation}

\noindent \textbf{State Embedding Stylization.}
The randomized styles $[\boldsymbol {\widetilde{\mu}_{i}},\boldsymbol{\widetilde{\sigma}}_{i}]^\mathrm{T}$ are injected into the original state embedding by AdaIN \cite{huang2017arbitrary}, computed as
\begin{equation} 
\label{AdaIN}
\boldsymbol{\widetilde{f}}_{i} =  \boldsymbol{\widetilde{\sigma}}_{i} \cdot \frac{\boldsymbol{f}_{i}-\boldsymbol{\mu}_{i}}{\boldsymbol{\sigma}_{i}}+ \boldsymbol{\widetilde{\mu}}_{i}.
\end{equation}

\subsection{State Flow Encoding}
\label{sec4.2}

After enriching the state style, 
how to learn a state embedding invariant to pre- and post- hallucinated styles is the key bottleneck. 
To address this, we introduce flow factorization \cite{song2023latent,song2024flow}, a recent representation disentanglement tool, to model the state embedding probability from each style as a flow by the gradient field in the latent space.
The mechanical dynamic optimal transport of the factorized flows allows the state embedding to be invariant to the styles.

\noindent \textbf{Generating Flow Embeddings.} 
Both $\boldsymbol{f}_{i}$ and $\boldsymbol{\widetilde{f}}_{i}$ are mapped to a latent embedding $\boldsymbol{z}_i$ and $\boldsymbol{\widetilde{z}}_i$ in the latent flow space by a Variational Auto-Encoder (VAE).
The architecture of VAE, for simplicity, directly follows \cite{song2024flow}.

\noindent \textbf{Prior State Flow Factorization.} 
The prior flow maps the transformation from the original state embedding $\boldsymbol{z}_i$ to the hallucinated state embedding $\boldsymbol{\widetilde{z}}_i$. 
Assume there are a total of $T$ steps in the factorization, the prior state flow 
can be factorized to $T$ terms, given by
\begin{equation} 
\label{FFpitoken}
p(\boldsymbol{z}_i,\boldsymbol{\widetilde{z}}_i)=p(\boldsymbol{\widetilde{z}}_{i,0})p(\boldsymbol{z}_{i,0}|\boldsymbol{\widetilde{z}}_{i,0})\prod \limits_{t=1}^{T}
p(\boldsymbol{\widetilde{z}}_{i,t}|\boldsymbol{\widetilde{z}}_{i,t-1})
p(\boldsymbol{z}_{i,t}|\boldsymbol{\widetilde{z}}_{i,t}).
\end{equation}

\noindent \textbf{Prior State Flow Evolution.} 
The prior flow evolves from the original state embedding to the hallucinated state embedding, which allows the probability density of the flow to be defined by the factorization. 
The conditional update $p(\boldsymbol{\widetilde{z}}_{i,t}|\boldsymbol{\widetilde{z}}_{i,t-1})$ is computed under a continuity equation form, given by $\partial_tp(\widetilde{z}_i)=- \nabla \cdot (p(\widetilde{z}_i) \nabla \psi (\widetilde{z}_i))$.
Here, $\nabla \psi (\widetilde{z}_i)$ denotes the induced velocity field, which is adverted by the potential function $\psi$ on the probability density $p(\widetilde{z}_i)$.
As the discrete particle evolution on the density is modeled as 
$\boldsymbol{\widetilde{z}}_{i,t} = f(\boldsymbol{\widetilde{z}}_{i,t-1})=\boldsymbol{\widetilde{z}}_{i,t-1}+\nabla_{\boldsymbol{\widetilde{z}}} \psi (\boldsymbol{\widetilde{z}}_{i,t-1})$, the conditional update of the prior flow evolution is computed as
\begin{equation} 
\label{PFE}
p(\boldsymbol{\widetilde{z}}_{i,t}|\boldsymbol{\widetilde{z}}_{i,t-1})=p(\boldsymbol{\widetilde{z}}_{i,t-1}) |\frac{d f(\boldsymbol{\widetilde{z}}_{i,t-1})}{d \boldsymbol{\widetilde{z}}_{i,t}}|^{-1}.
\end{equation}

The diffusion equation $\psi = - D {\rm log}p(\boldsymbol{\widetilde{z}}_{i,t})$, which represents the random trajectories with a minimum of informative prior, allows the prior flow to be evolved as
\begin{equation} 
\label{PFE2}
\partial_tp(\boldsymbol{\widetilde{z}}_{i,t})=- \nabla \cdot (p(\boldsymbol{\widetilde{z}}_{i,t}) \nabla \psi) = D \nabla^2 p(\boldsymbol{\widetilde{z}}_{i,t}),
\end{equation}
where $D$ is a constant coefficient.

\noindent \textbf{Posterior State Factorization.} 
In contrast to the prior flow, the posterior flow maps the approximation from the latent embedding of the hallucinated state $\boldsymbol{\widetilde{z}}_i$ to the latent embedding of the original state $\boldsymbol{z}_i$.
The posterior flow can be factorize as
\begin{equation} 
\label{FFpotoken}
q(\boldsymbol{\widetilde{z}}_i|\boldsymbol{z}_i)=q(\boldsymbol{\widetilde{z}}_{i,0}|\boldsymbol{z}_{i,0})\prod \limits_{t=1}^{T}
q(\boldsymbol{\widetilde{z}}_{i,t}|\boldsymbol{\widetilde{z}}_{i,t-1}).
\end{equation}

\noindent \textbf{Posterior State Evolution.} 
Same as the prior flow evolution of the token features, the continuity equation is used to model the posterior flow evolution. 
Specifically, given the 
particle evolution function
$\boldsymbol{\widetilde{z}}_{i,t} = g(\boldsymbol{\widetilde{z}}_{i,t-1})=\boldsymbol{\widetilde{z}}_{i,t-1}+\nabla_{\boldsymbol{\widetilde{z}}} u$, it is mathematically computed as
\begin{equation} 
\label{PoFE}
q(\boldsymbol{\widetilde{z}}_{i,t}|\boldsymbol{\widetilde{z}}_{i,t-1})=q(\boldsymbol{\widetilde{z}}_{i,t-1}) |\frac{d g(\boldsymbol{\widetilde{z}}_{i,t-1})}{d \boldsymbol{\widetilde{z}}_{i,t-1}}|^{-1}.
\end{equation}

After discretizing the above equation and implementing the logarithm operation, Eq.~\ref{PoFE}~can be mathematically re-formulated as
\begin{equation} 
\label{PoFE2}
{\rm log} q(\boldsymbol{\widetilde{z}}_{i,t}|\boldsymbol{\widetilde{z}}_{i,t-1})={\rm log} q(\boldsymbol{\widetilde{z}}_{i,t-1}) - {\rm log} |1+\nabla_{\boldsymbol{\widetilde{z}}}^{2} u|.
\end{equation}

\subsection{State Flow Constraint}
\label{sec4.3}

After modeling the original state embedding and the hallucinated state embedding by the prior and posterior flow, it is necessary to constrain them in the latent flow space, so that both state embeddings are enforced to learn the same representation despite style variance.
In flow factorization \cite{song2023latent,song2024flow}, this constraint is realized by the latent posterior with the  
optimal transport path.

\noindent \textbf{Definition 1. Benamou-Brenier Formula.} 
\textit{Given two probability measures $\mu_0$ and $\mu_1$, their $L2$ Wasserstein distance is defined by}
\begin{equation} 
\label{W2}
W_2(\mu_0,\mu_1)^2=\mathop{{\rm min}}\limits_{\rho,\nu}\{\int \int \frac{1}{2} \rho(x,t) |\nu(x,t)|^2 dx dt \}.
\end{equation}
\textit{where the density $\rho$ and the velocity $\nu$ satisfy:}
\begin{equation} 
\label{pv}
\frac{d\rho(x,t)}{dt}= - \nabla \cdot (\nu(x,t)\rho(x,t)), \nu(x,t)=\nabla u(x,t).
\end{equation}

Specifically, when $\nabla u$ satisfies certain partial differential equation (PDE), the probability density evolution between the original state embedding and the hallucinated state embedding can be minimized by 
the $L2$ Wasserstein distance. 
The generalized Hamilton-Jacobi (HJ) equation
(i.e., $\partial_t u + 1/2 ||\nabla u||^2 \leq 0$)
determines the optimality condition of the velocity.
Consequently, the posterior flow of the state embedding is supposed to satisfy the HJ equation with an external driving force, given by
\begin{equation} 
\label{HJ}
\frac{\partial}{\partial t} u (\boldsymbol{\widetilde{z}}_{i},t) + \frac{1}{2} ||\nabla_{\boldsymbol{\widetilde{z}}_{i}} u(\boldsymbol{\widetilde{z}}_{i},t)||^2 = f(\boldsymbol{\widetilde{z}}_{i},t) \quad s.t.  f(\boldsymbol{\widetilde{z}}_{i},t) \leq 0.
\end{equation}

To realize the negative constraint of this external force $f(\boldsymbol{\widetilde{z}}_{i},t)$, a MLP is used for parameterization, given by 
$f(\boldsymbol{\widetilde{z}}_{i},t) = - {\rm MLP}([\boldsymbol{\widetilde{z}}_{i};t])^2$. 
For simplicity, the MLP we use for $f(\boldsymbol{\widetilde{z}}_{i},t)$ shares the same architecture as $u(\boldsymbol{\widetilde{z}}_{i},t)$ does. 
The above PDE constraint is realized by a physics-informed neural network loss \cite{raissi2019physics}, given by 
\begin{footnotesize}
\begin{equation} 
\begin{aligned}
\label{PINNloss}
\mathcal{L}_{HJ}^{i} & = \frac{1}{T} \sum_{t=1}^{T} (\frac{\partial}{\partial t} u (\boldsymbol{\widetilde{z}}_{i},t) + \frac{1}{2} ||\nabla_{\boldsymbol{\widetilde{z}}_{i}} u(\boldsymbol{\widetilde{z}}_{i},t)||^2 \\
& - f(\boldsymbol{\widetilde{z}}_{i},t))^{2} +||\nabla u(\boldsymbol{\widetilde{z}}_0,0)||^2,
\end{aligned}
\end{equation}
\end{footnotesize}
where the first term allows the flow to be constrained by the HJ equation, and the second term matches the initial condition.
This constraint allows the posterior flow from the original state embedding and the hallucinated state embedding to be optimally aligned, so that the impact caused by the style variation is eliminated.  

\subsection{Implementation Details}

The proposed DGFamba uses VMamba \cite{liu2024vmamba} as the backbone.
The initial weights of VMamba have been pre-trained on ImageNet \cite{deng2009imagenet}. 
The image encoder consists of four blocks, with a number of 2, 2, 4 and 2 VMamba layers.
The proposed three key steps are integrated into each of these VMamba layers.
The total loss $\mathcal{L}$ is a linear combination between the classification loss $\mathcal{L}_{cls}$ and the HJ loss $\mathcal{L}_{HJ}$ defined in Eq.~\ref{PINNloss}, given by $
\mathcal{L} = \mathcal{L}_{cls} + \sum_{i=1}^{N} \mathcal{L}_{HJ}^i$.

For fair evaluation with DGMamba~\cite{long2024dgmamba}, the configuration settings keep the same.
The training terminates after 10000 iterations, with a batch size of 16 per source domain.
The AdamW optimizer is used for optimization, with a momentum value of 0.9 and an initial learning rate of $3\times10^{-4}$.
In addition, the cosine decay learning rate scheduler is adapted.

\begin{table*}[!t]
    \centering
    \begin{tabular}{c|c|c|cccc|c}
\hline \multirow{2}{*}{ Method } & \multirow{2}{*}{Venue } & \multirow{2}{*}{ Params. } & \multicolumn{4}{|c|}{ Target domain } & \multirow{2}{*}{ Avg.( $\uparrow$ ) } \\
\cline { 4 - 7 } & & & Art & Cartoon & Photo & Sketch & \\
\hline 
\textit{ResNet-50 Based:} & \\
GroupDRO & ICLR 2019 & 23M & 83.5 & 79.1 & 96.7 & 78.3 & 84.4 \\
VREx & ICML 2021 & 23M & 86.0 & 79.1 & 96.9 & 77.7 & 84.9 \\
RSC & ECCV 2020 & 23M & 85.4 & 79.7 & 97.6 & 78.2 & 85.2 \\
MTL & JMLR 2021 & 23M & 87.5 & 77.1 & 96.4 & 77.3 & 84.6 \\
Mixstyle & ICLR 2021 & 23M & 86.8 & 79.0 & 96.6 & 78.5 & 85.2 \\
SagNet & CVPR 2021 & 23M & 87.4 & 80.7 & 97.1 & 80.0 & 86.3 \\
ARM & NeurIPS 2021 & 23M & 86.8 & 76.8 & 97.4 & 79.3 & 85.1 \\
SWAD & NeurIPS 2021 & 23M & 89.3 & 83.4 & 97.3 & 82.5 & 88.1 \\
PCL & CVPR 2022 & 23M & 90.2 & 83.9 & 98.1 & 82.6 & 88.7 \\
SAGM & CVPR 2023 & 23M & 87.4 & 80.2 & 98.0 & 80.8 & 86.6 \\
iDAG & ICCV 2023 & 23M & 90.8 & 83.7 & 98.0 & 82.7 & 88.8 \\
GMDG & CVPR 2024 & 23M & 84.7 & 81.7 & 97.5 & 80.5 & 85.6 \\
\hline 
\textit{DeiT-S Based:} & \\
SDViT & ACCV 2022 & 22M & 87.6 & 82.4 & 98.0 & 77.2 & 86.3 \\
GMoE & ICLR 2023 & 34M & 89.4 & 83.9 & 99.1 & 74.5 & 86.7 \\
\hline 
\textit{VMamba Based:} & \\
DGMamba & MM 2024 & 31M & 91.3 & 87.0 & 99.0 & 87.3 & 91.2 \\
\textbf{DGFamba} & AAAI 2025 & 31M & \textbf{92.6} & \textbf{89.4} & \textbf{99.7} & \textbf{88.8} & \textbf{92.6} \\
\hline
\end{tabular}
 \caption{
 Performance comparison between the proposed DGFamba and existing methods on PACS dataset. M: in million.
 }
\label{resPACS}
\end{table*}

\begin{table}[!t]
    \centering
    \resizebox{\linewidth}{!}{
    \begin{tabular}{c|cccc|c}
\hline \multirow{2}{*}{ Method } & \multicolumn{4}{|c|}{ Target domain } & \multirow{2}{*}{ Avg.( $\uparrow$ ) } \\
\cline { 2 - 5 } & C & L & S & P & \\
\hline 
\textit{ResNet-50 Based:} & \\
GroupDRO & 97.3 & 63.4 & 69.5 & 76.7 & 76.7 \\
VREx & 98.4 & 64.4 & 74.1 & 76.2 & 78.3 \\
RSC & 97.9 & 62.5 & 72.3 & 75.6 & 77.1 \\
MTL & 97.8 & 64.3 & 71.5 & 75.3 & 77.2 \\
Mixstyle & 98.6 & 64.5 & 72.6 & 75.7 & 77.9 \\
SagNet & 97.9 & 64.5 & 71.4 & 77.5 & 77.8 \\
ARM & 98.7 & 63.6 & 71.3 & 76.7 & 77.6 \\
SWAD & 98.8 & 63.3 & 75.3 & 79.2 & 79.1 \\
PCL & 99.0 & 63.6 & 73.8 & 75.6 & 78.0 \\
SAGM & 99.0 & 65.2 & 75.1 & 80.7 & 80.0 \\
iDAG & 98.1 & 62.7 & 69.9 & 77.1 & 76.9 \\
GMDG & 98.3 & 65.9 & 73.4 & 79.3 & 79.2 \\
\hline 
\textit{DeiT-S Based:} & \\
SDViT & 96.8 & 64.2 & 76.2 & 78.5 & 78.9 \\
GMoE & 96.9 & 63.2 & 72.3 & 79.5 & 78.0 \\
\hline 
\textit{VMamba Based:} & \\
DGMamba & 98.9 & 64.3 & 79.2 & 80.8 & 80.8 \\
\textbf{DGFamba} & \textbf{99.5} & \textbf{66.2} & \textbf{80.9} & \textbf{82.0} & \textbf{82.2} \\
\hline
\end{tabular}
}
\caption{Performance comparison between the proposed DGFamba and existing state-of-the-art methods on VLCS dataset. 
C: Caltech; L: LabelMe; S: SUN; P: PASCAL.}
\label{resVLCS}
\end{table}

\begin{table}[!t]
\centering
\resizebox{\linewidth}{!}{
\begin{tabular}{c|cccc|c}
\hline 
\multirow{2}{*}{ Method } & \multicolumn{4}{|c|}{ Target domain } & \multirow{2}{*}{ Avg. $(\uparrow)$} \\
\cline { 2 - 5 } & A & C & P & R & \\
\hline 
\textit{ResNet-50 Based:} & \\
GroupDRO & 60.4 & 52.7 & 75.0 & 76.0 & 66.0 \\
VREx & 60.7 & 53.0 & 75.3 & 76.6 & 66.4 \\
RSC & 60.7 & 51.4 & 74.8 & 75.1 & 65.5 \\
MTL & 61.5 & 52.4 & 74.9 & 76.8 & 66.4 \\
Mixstyle & 51.1 & 53.2 & 68.2 & 69.2 & 60.4 \\
SagNet & 63.4 & 54.8 & 75.8 & 78.3 & 68.1 \\
ARM & 58.9 & 51.0 & 74.1 & 75.2 & 64.8 \\
SWAD & 66.1 & 57.7 & 78.4 & 80.2 & 70.6 \\
PCL & 67.3 & 59.9 & 78.7 & 80.7 & 71.6 \\
SAGM & 65.4 & 57.0 & 78.0 & 80.0 & 70.1 \\
iDAG & 68.2 & 57.9 & 79.7 & 81.4 & 71.8 \\
GMDG & 68.9 & 56.2 & 79.9 & 82.0 & 70.7 \\
\hline 
\textit{DeiT-S Based:} & \\
SDViT & 68.3 & 56.3 & 79.5 & 81.8 & 71.5 \\
GMoE & 69.3 & 58.0 & 79.8 & 82.6 & 72.4 \\
\hline 
\textit{VMamba Based:} & \\
DGMamba & 76.2 & 61.8 & 83.9 & 86.1 & 77.0 \\
\textbf{DGFamba} & \textbf{77.4} & \textbf{63.7} & \textbf{85.6} & \textbf{87.3} & \textbf{78.5} \\
\hline
\end{tabular}
}
\caption{Performance comparison between the proposed DGFamba and existing state-of-the-art methods on OfficeHome. 
A: Art; C: Clipart; P: Product; R: Real.}
\label{resOff}
\end{table}

\section{Experiments}
\label{Sec4}

\subsection{Datasets \& Evaluation Protocols}

\textbf{Datasets.} Our experiments are conducted on four visual domain generalization datasets.
Specifically, \textbf{PACS}~\cite{li2017deeper}~
consists of 9,991 samples from four domains. 
\textbf{VLCS}~\cite{fang2013unbiased}~consists of a total number of 10,729 samples from four domains.
\textbf{OfficeHome}~\cite{venkateswara2017deep}~consists of 15,588 samples from four different domains. 
\textbf{TerraIncognita}~\cite{beery2018recognition}~consists of 24,330 samples from four different domains. 

\noindent\textbf{Evaluation Protocols.} 
Following the evaluation protocols of existing methods~\cite{gulrajani2020search,cha2021swad}, experiments are conducted under the leave-one-domain-out protocol, where only one domain is used as the unseen target domain and the rest domains are used as the source domains for training. 
The classification accuracy (in percentage, \%) is used as the evaluation metric. 

\subsection{Comparison with State-of-the-art}

Existing visual domain generalization methods are compared.
The first category is CNN based methods, including GroupDRO~\cite{sagawa2019distributionally}, VREx~\cite{krueger2021out}, RSC~\cite{huang2020self}, MTL~\cite{blanchard2021domain}, Mixstyle~\cite{zhou2024mixstyle}, SagNet~\cite{nam2021reducing}, ARM~\cite{blanchard2021domain}, SWAD~\cite{cha2021swad}, PCL~\cite{yao2022pcl}, SAGM~\cite{wang2023closer}, iDAG~\cite{huang2023idag}, and GMDG~\cite{tan2024rethinking}.
The second category is ViT based methods, including SDViT~\cite{sultana2022self} and GMoE~\cite{li2023sparse}.
The third category is the contemporary VMamba based method, namely, DGMamba~\cite{long2024dgmamba}.

\noindent \textbf{Results on PACS} are reported in
Table~\ref{resPACS}. 
DGFamba shows the best performance on all the four experimental settings, yielding an accuracy of 92.6\%, 89.4\%, 99.7\% and 88.8\% on A, C, P and S unseen target domain, respectively. 
The average accuracy achieves 92.6\%, outperforming the second-best DGMamba by 1.4\%.
Specifically, the accuracy improvement on C and S unseen target domains is 2.4\% and 1.5\%, respectively.
It also significantly outperforms existing CNN and ViT based methods by an improvement about 6\%, while at the same time has less parameter number.

\noindent \textbf{Results on VLCS} are reported in
Table~\ref{resVLCS}. 
DGFamba outperforms all the compared methods under all the four experiment settings. 
It outperforms the second-best, DGMamba, by 1.4\% average accuracy.
Notably, the accuracy improvement on unseen L and S  domains is 1.9\% and 1.7\%, respectively. 
These outcomes indicate that the proposed DGFamba is more stable and more robust when generalized on unseen target domains. 
On the other hand, DGFamba outperforms the best CNN based method SAGM by an average accuracy of 2.2\%, and outperforms the best ViT based method SDViT by an average accuracy of 3.3\%.

\noindent \textbf{Results on OfficeHome} are reported in 
Table~\ref{resOff}.
DGFamba shows state-of-the-art performance on all the four unseen target domains, yielding an average accuracy of 78.5\%.
It significantly outperforms the second-best DGMamba.
The accuracy improvement on the A, C, P and R unseen target domain is 1.2\%, 1.9\%, 1.7\% and 1.2\%, respectively. 
DGMamba outperforms the best CNN based method PCL by an average accuracy of 6.9\%, and surpasses the best ViT based method GMoE by an average accuracy of 6.1\%.

\noindent \textbf{Results on TerraIncognita.} 
Table~\ref{resTerra}~compares the performance. 
Same as the above three experiments, the proposed DGFamba outperforms all existing methods on all the four unseen target domains, yielding an average accuracy of 56.1\%.
Compared with the second-best, the accuracy improvement on the L100, L38, L43 and L46 unseen target domain is 1.2\%, 1.5\%, 2.0\% and 1.1\%, respectively. 
Compared with the best-performed CNN based method PCL, the average accuracy improvement is 4.0\%,
Compared with the best-performed ViT based method GMoE, the average accuracy improvement is 10.5\%.

\subsection{Ablation Studies}

\noindent \textbf{On Each Component.}
On top of the VMamba backbone, the proposed DGFamba consists of three key components, namely, State Style Randomization (SSR), State Flow Encoding (SFE), and State Flow Constraint (SFC), respectively.
When there is no SFC component, the feature representation processed by SSR or SFE is processed by a MLP to finish the feature propagation.
Table~\ref{ablcon}~inspects the performance of each individual component. 
SSR mainly focuses on enriching the style diversity. Naively using it functions as a type of feature augmentation, which leads to an average accuracy improvement of 1.0\%.
Similarly, SFE helps further condense the feature embedding from both the pre- and post- style randomized samples in the latent flow space.
It also leads an average accuracy improvement of 1.1\%.
Finally, SFC constrains the feature distribution between the pre- and post- style randomized samples in the latent flow space, which contributes to the most significant improvement.
Especially, the accuracy improvement on the A, C, P and S unseen target domain is 1.9\%, 1.2\%, 0.6\% and 0.9\%. 

\begin{table}[!t]
    \centering
    \resizebox{\linewidth}{!}{
    \begin{tabular}{c|cccc|c}
\hline \multirow{2}{*}{ Method } & \multicolumn{4}{|c|}{ Target domain } & \multirow{2}{*}{ Avg. $(\uparrow)$} \\
\cline{2-5} & L100 & L38 & L43 & L46 & \\
\hline 
\textit{ResNet-50 Based:} & \\
GroupDRO & 41.2 & 38.6 & 56.7 & 36.4 & 43.2 \\
VREx & 48.2 & 41.7 & 56.8 & 38.7 & 46.4 \\
RSC & 50.2 & 39.2 & 56.3 & 40.8 & 46.6 \\
MTL & 49.3 & 39.6 & 55.6 & 37.8 & 45.6 \\
Mixstyle & 54.3 & 34.1 & 55.9 & 31.7 & 44.0 \\
SagNet & 53.0 & 43.0 & 57.9 & 40.4 & 48.6 \\
ARM & 49.3 & 38.3 & 55.8 & 38.7 & 45.5 \\
SWAD & 55.4 & 44.9 & 59.7 & 39.9 & 50.0 \\
PCL & 58.7 & 46.3 & 60.0 & 43.6 & 52.1 \\
SAGM & 54.8 & 41.4 & 57.7 & 41.3 & 48.8 \\
iDAG & 58.7 & 35.1 & 57.5 & 33.0 & 46.1 \\
GMDG & 59.8 & 45.3 & 57.1 & 38.2 & 50.1 \\
\hline
\textit{DeiT-S Based:} & \\
SDViT & 55.9 & 31.7 & 52.2 & 37.4 & 44.3 \\
GMoE & 59.2 & 34.0 & 50.7 & 38.5 & 45.6 \\
\hline
\textit{VMamba Based:} & \\
DGMamba & 62.7 & 48.3 & 61.1 & 46.4 & 54.6 \\
\textbf{DGFamba} & \textbf{63.9} & \textbf{49.8} & \textbf{63.1} & \textbf{47.5} & \textbf{56.1} \\
\hline
\end{tabular}
}
\caption{Performance comparison between the proposed DGFamba and existing state-of-the-art methods on TerraIncognita. The best results are marked in \textbf{bold}.}
\label{resTerra}
\end{table}

\begin{table}[!t]
    \centering
    \resizebox{\linewidth}{!}{
    \begin{tabular}{ccc|cccc|c}
\hline \multicolumn{3}{c|}{Component} & \multicolumn{4}{|c|}{ Target domain } & \multirow{2}{*}{ Avg. $\uparrow$} \\
\cline{1-7}
SSR & SFE & SFC & Art & Cartoon & Photo & Sketch & \\
\hline 
$\usym{2717}$ & $\usym{2717}$ & $\usym{2717}$ & 88.2 & 86.2 & 98.4 & 84.9 & 89.4 \\
$\checkmark$ & $\usym{2717}$ & $\usym{2717}$ & 89.4 & 87.1 & 98.7 & 86.2 & 90.4 \\
$\checkmark$ & $\checkmark$ & $\usym{2717}$ & 90.7 & 88.2 & 99.1 & 87.9 & 91.5 \\
$\checkmark$ & $\checkmark$ & $\checkmark$ & \textbf{92.6} & \textbf{89.4} & 
\textbf{99.7} & \textbf{88.8} & \textbf{92.6} \\
\hline
\end{tabular}
}
\caption{Ablation studies on each component in the proposed DGFamba. VMamba \cite{liu2024vmamba} as baseline.
Experiments are conducted on the PACS dataset. 
}
\label{ablcon}
\end{table}

\noindent \textbf{On Each Feature Block.}
As the proposed DGFamba implements the flow factorization in every layer, it is also necessary to inspect the contribution to generalization from different layers.
To this end, Table~\ref{ablblock}~ablates the block-wise contribution, where $\mathbf{F}_1$, $\mathbf{F}_2$, $\mathbf{F}_3$ and $\mathbf{F}_4$ denote the implementation on the first, second, third and fourth VMamba block. 

It can be seen that, implementing the proposed learning scheme shows a more predominant performance improvement on unseen target domains. 
Specifically, implementing on the first VMamba block ($\mathbf{F}_1$) can lead to an improvement of 1.5\%, 1.3\%, 0.5\% and 1.2\% on the Art, Cartoon, Photo and Sketch unseen target domain, respectively.
This may be explained that the shallower features are usually more sensitive to the shift of color, shape and etc, which are typical factors of the cross-domain style.
In contrast, the deeper features usually rest more semantic and content information, which is less sensitive to the shift of cross-domain styles. 
Nevertheless, implementing on the fourth VMamba block ($\mathbf{F}_4$) still exhibits a clear accuracy improvement on unseen domains, namely, 0.7\% on Art, 0.4\% on Cartoon, 0.2\% on Photo and 0.6\% on Sketch, respectively. 

\begin{table}[!t]
    \centering
    \resizebox{\linewidth}{!}{
    \begin{tabular}{cccc|cccc}
\hline \multicolumn{4}{c|}{Feature Block} & \multicolumn{4}{|c}{ Target domain } \\
\cline{1-8}
$\mathbf{F}_1$ & $\mathbf{F}_2$ & $\mathbf{F}_3$ & $\mathbf{F}_4$ & Art & Cartoon & Photo & Sketch \\
\hline 
$\usym{2717}$ & $\usym{2717}$ & $\usym{2717}$ & $\usym{2717}$ & 88.2 & 86.2 & 98.4 & 84.9 \\
$\checkmark$ & $\usym{2717}$ & $\usym{2717}$ & $\usym{2717}$ & 89.7 & 87.5 & 98.9 & 86.1 \\
$\checkmark$ & $\checkmark$ & $\usym{2717}$ & $\usym{2717}$ & 91.0 & 88.6 & 99.3 & 87.3 \\
$\checkmark$ & $\checkmark$ & $\checkmark$ & $\usym{2717}$ & 91.9 & 89.0 & 99.5 & 88.2\\
$\checkmark$ & $\checkmark$ & $\checkmark$ & $\checkmark$ & \textbf{92.6} & \textbf{89.4} & 
\textbf{99.7} & \textbf{88.8} \\
\hline
\end{tabular}
}
\caption{Ablation studies on each component in the proposed DGFamba. VMamba \cite{liu2024vmamba} as baseline.
Experiments are conducted on the PACS dataset. 
}
\label{ablblock}
\end{table}

\noindent \textbf{On Factorization Steps}
$T$ manipulates the factorization steps in the probability latent space.
By default the factorization step $T$ is set to 8.
We further test the results when $T$ shifts from 2 to 12, with an interval of 2.
The results reported in Table~\ref{ablT} show that,  the best performance on unseen domains is achieved when $T$ is around 6 or 8.
A factorization step that is too small or too large leads to a decline in performance, which may be explained that a small/large factorization step $T$ can under-/over- fit the probability path, negatively impacting the overall generalization ability. 

\begin{table}[!t]
    \centering
    \begin{tabular}{c|cccc|c}
\hline 
T & Art & Cartoon & Photo & Sketch & avg. \\
\hline 
2 & 91.5 & 88.6 & 98.8 & 87.6 & 91.6 \\
4 & 91.9 & 89.0 & 99.2 & 88.1 & 92.1 \\
6 & 92.3 & 89.2 & 99.5 & 88.4 & 92.4 \\
8 & \textbf{92.6} & \textbf{89.4} & \textbf{99.7} & \textbf{88.8} & \textbf{92.6} \\
10 & 92.5 & 89.3 & 99.4 & 88.5 & 92.4 \\
12 & 92.1 & 88.7 & 99.0 & 88.2 & 92.0 \\
\hline
\end{tabular}
   \caption{Impact of the factorization step $T$ on generalization performance. Experiments are conducted on PACS dataset.}
    \label{ablT}
\end{table}

\begin{figure}[!t]
  \centering
   \includegraphics[width=1.0\linewidth]{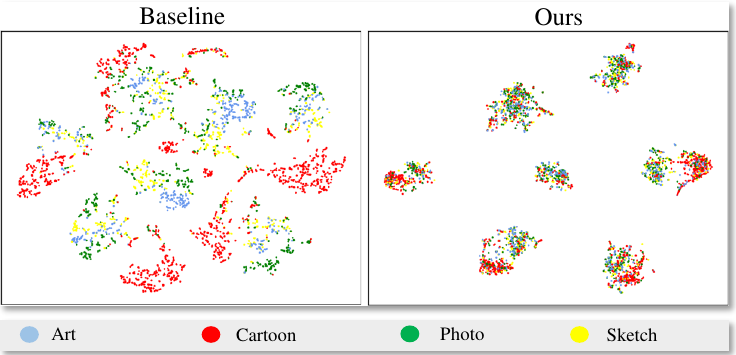}
   \vspace{-0.5cm}
   \caption{t-SNE feature space visualization between the Mamba baseline and the proposed DGFamba. Experiments conducted on the PACS dataset. A more generalized representation allows features from different domains (in different color) to be uniformly distributed. 
   }
   \label{vis}
\end{figure}

\begin{figure}[!t]
  \centering
   \includegraphics[width=1.0\linewidth]{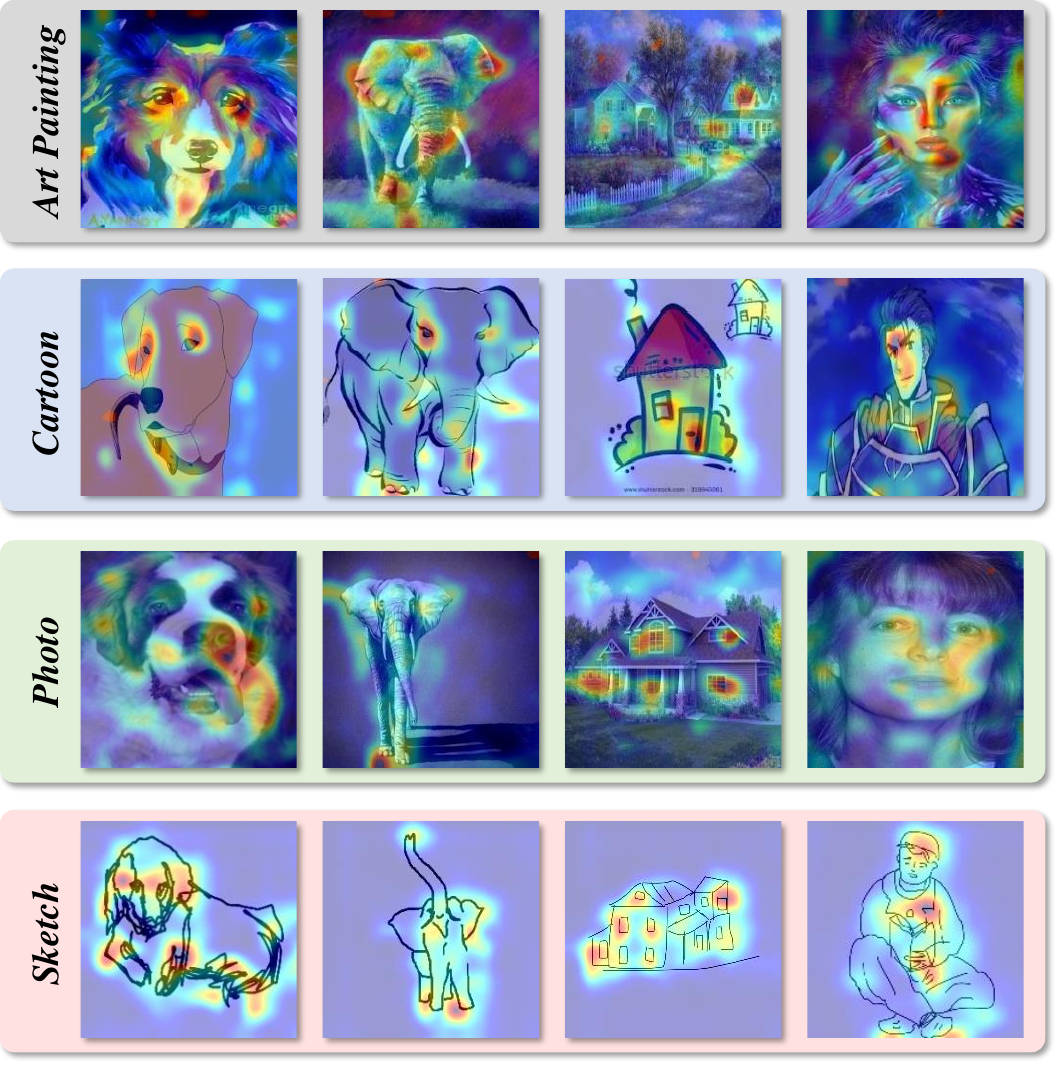}
   \vspace{-0.5cm}
   \caption{Attention map on the unseen target domain from four experiments on PACS.
   }
   \label{heatmap}
\end{figure}

\subsection{Understanding Flow Factorized State Space}

\noindent \textbf{t-SNE visualization.}
We extract the feature embedding of the VMamba baseline and the proposed DGFamba, and inspect their distribution by t-SNE visualization.
The results are displayed in Fig.~\ref{vis}a and b, respectively.
The samples from different domains in the PACS dataset are labeled in different types of color.
Ideally, a generalized representation allows the feature embeddings from different domains (in different color) to be more uniformly distributed, which corresponds to the observation in Fig.~\ref{vis}.
This further indicates the generalization ability of DGFamba. 

\noindent \textbf{Activation Map Visualization.}
Fig.~\ref{heatmap}~visualizes the attention map of the proposed DGFamba on the unseen target domain across all four experimental settings.
We use the class activation map (CAM) \cite{zhou2016learning} to compute the heat map and layout on the original image.
It can be seen that DGMamba highlights the key local regions of each category, despite the style shift on unseen target domain.

\section{Conclusion}

In this paper, we proposed a flow factorized state space learning scheme to harness the emerging selective state space modeling (SSM) for visual domain generalization. 
Its general idea is to learn a style-invariant state space embedding by first randomizing the styles and then aligning the pre- and post- hallucinated state embedding in the latent flow space.
The proposed DGFamba consists of three key components, namely, state style randomization (SSR), state flow encoding (SFE) and state flow constraint (SFC).
Extensive experiments demonstrated that DGFamba significantly outperforms existing CNN and ViT based methods, and the concurrent DGMamba.

\bibliography{aaai25}

\end{document}